\pdfoutput=1
\documentclass[11pt]{article}

\usepackage{ACL2023}
\usepackage{times}
\usepackage{latexsym}
\usepackage{todonotes}
\usepackage{float}
\usepackage{tablefootnote}
\usepackage[normalem]{ulem}
\usepackage[T1]{fontenc}
\usepackage[utf8]{inputenc}

\usepackage{microtype}

\usepackage{inconsolata}

\title{Simple Augmentations of Logical Rules for \\ Neuro-Symbolic Knowledge Graph Completion}
\author{Ananjan Nandi \hskip 1em  
Navdeep Kaur  \hskip 1em
Parag Singla \hskip 1em Mausam \\
  Indian Institute of Technology, Delhi \\
  \texttt{\{tgk.ananjan, navdeepkjohal\}@gmail.com} \quad 
   \texttt{\{parags, mausam\}@cse.iitd.ac.in}  
}

\begin{document}
\maketitle
\begin{abstract}
High-quality and high-coverage rule sets are imperative to the success of Neuro-Symbolic Knowledge Graph Completion (NS-KGC) models, because they form the basis of all symbolic inferences. Recent literature builds neural models for generating rule sets, however, preliminary experiments show that they struggle with maintaining high coverage. In this work, we suggest three simple augmentations to existing rule sets: (1) transforming rules to their abductive forms, (2) generating equivalent rules that use inverse forms of constituent relations and (3) random walks that propose new rules. Finally, we prune potentially low quality rules. Experiments over four datasets and five ruleset-baseline settings suggest that these simple augmentations consistently improve results, and obtain up to 7.1 pt MRR and 8.5 pt Hits@1 gains over using rules without augmentations. 

%High-quality rule sets are imperative to the success of Neuro-Symbolic Knowledge Graph Completion (NS-KGC) models, because they drive the performance of the model by encoding relevant information about the dataset. Learning logical rules from a KG is an inherently hard task. The past neuro-symbolic models have struggled to learn the amount of high-quality rules required for large or dense KGs. To address these limitations, we propose three novel ways of feeding more high-quality rules to Neuro-Symbolic KG models. Specifically, we augment an existing set of learned rules by exploiting the concepts of abduction and rule inversion. Further, we learn new rules by using PCA-scores over local random walks. Experiments on four datasets prove that our proposed approach obtains up to 18\% Hits@1 and 12\% MRR gains over baselines. 
%Further analysis reveals that [one line conclusion from the ablation study done on RotatE vs Abduction]
\end{abstract}

\section{Introduction}
\label{sec:introduction}
Knowledge Graphs (KGs) comprise important world knowledge facts, but are typically incomplete, 
due to their ever-increasing size. KG embeddings ~\cite{KGESurvey2017} has been the dominant methodology for knowledge graph completion (KGC). 
A KG embedding approach represents entities and relations as learnable dense vectors and computes a score for an unseen fact as a function over them. These generally have state-of-the-art performance, especially for large KGs.

Recently, neuro-symbolic (NS-KGC) approaches for the task  have been proposed, where KG embeddings are enhanced by inferences over an explicit first-order logic rule set ~\cite{ExpressGNN2020, RNNLogic2021}. The resulting models bring together best of both worlds -- generalizability and interpretability of explicit logical rules, and the scalability and representation power of embeddings. Unfortunately, a key roadblock for success of NS-KGC is the availability of a high-coverage rule set. 

%the research focus has shifted towards designing advanced Neuro-Symbolic (NS) models~\cite{ExpressGNN2020, RNNLogic2021} for knowledge graph completion where KG embeddings are enhanced by learning first-order logic rules along with them. The resulting models bring together best of both the worlds - generalizability and interpretability of first-order logic rules with the scalability and superior performance of KG embeddings. 

Early NS-KGC methods, such as NeuralLP~\cite{NeuralLP2017} and DRUM~\cite{Drum2019}, learn rules as part of a single model, %\todo{M:check. also i removed RUGE and ExpressGNN from here. Discuss on phone} b
but do not have performance competitive with embedding models such as RotatE~\cite{sun2018rotate}. A recent NS-KGC model, RNNLogic \cite{RNNLogic2021}, matches empirical performance with embedding approaches. It has a separate neural component that outputs a set of rules, which is then used to train inference parameters, in an EM-based approach. Preliminary experiments on RNNLogic suggest that its ruleset has limited coverage, due to which symbolic inferences do not fire for many queries, and the model gets limited to using its embedding part only. The goal of this work is to strengthen the symbolic inferences in NS-KGC models for better overall performance.

In this work, we propose simple augmentations that take an existing ruleset (such as one output by RNNLogic) and proposes additional (related) rules to improve coverage and quality. We propose three augmentations. First, we convert each deductive rule into its abductive counterparts. Second, we supplement each rule via an equivalent rule that uses inverses for all constituent relations. Third, we generate additional high-quality rules independently by local random walks and subsequent PCA filtering~\cite{AMIE2013}. These increase size of ruleset drastically; we balance runtimes by additionally pruning rules from existing set using our filtering approach. Overall, this results in a comparable number of high-quality and high-coverage rules, for use in NS-KGC.

%The core intention behind our proposed approach is to extract maximum knowledge from the learnt rules and assimilate it into supplementary rules. 
%Our proposed approach is generic and can be integrated with most of the existing NS-KG models. 

On four KGC datasets, over three NS-KGC models, we find that our augmentations consistently improve KGC performance, outperforming no augmentation baselines by up to 7.1 MRR and 8.5 Hits@1 pts. We believe our augmentations should become standard practice over any ruleset for NS-KGC. We release our code \footnote{\url{https://github.com/dair-iitd/NS-KGC-AUG}} and rulesets. 

%By performing experiments over four datasets, we show that our proposed augmentations outperform base model by significant margins while acquiring rules from different sources.

%To summarize, we make the following \textbf{contributions in this work}: 
%\begin{enumerate}
%\item We propose an effective rule augmentation technique for neuro-symbolic KG models. Specifically, (a) we explore the concept of abduction and rule inversion in NS-KG models in order to distill most knowledge from existing rules. 
%\item Our technique is generic and is applicable to any rule-based KG model. As a proof, we employ it on two diverse models, RNNLogic and RLogic. 
%\item Our model shows improvement by x points MRR on some datasets.
%\end{enumerate}

\section{Background and Related Work}
\label{sec:background and related work}
We are given an incomplete KG $\mathcal{K} = (\mathcal{E}, \mathcal{R}, \mathcal{T})$ consisting of entities $\mathcal{E}$, relation set $\mathcal{R}$ and 
set $\mathcal{T}=\{(\fol{h},\fol{r},\fol{t})\}$ of triples. Our goal is to predict the validity of any triple not present in $\mathcal{T}$. 

\vspace{0.5ex}
\noindent \textbf{Related Work:} 
\label{sec:related work}
Existing work on NS-KGC can roughly be characterized into four types. One approach is to use attention over relations to learn end-to-end differentiable models~\cite{NeuralLP2017, Drum2019}. A second approach, which includes Minerva~\cite{MINERVA2018} and DeepPath~\cite{deeppath2017}, uses RL to train an agent to find reasoning paths for KG completion. These approaches are not yet competitive to KG embedding models for large datasets. Thirdly, models like ExpressGNN \cite{ExpressGNN2020} and RNNLogic use variational inference to assess plausibility of a given triple. We experiment with both these models in this paper. The final type includes UNIKER~\cite{UNIKER2021} and RUGE~\cite{RUGE2018}, which integrate embeddings alongside traditional rules learnt via ILP models. We believe that our augmented rules can benefit these works too. Since our experiments are based on RNNLogic, ExpressGNN and we utilize PCA scores for filtering, we describe these in some detail next.

%The first axis including RNNLogic~(\citeyear{RNNLogic2021}), ExpressGNN~(\citeyear{ExpressGNN2020}) exploit variational inference to assess the plausibility of a given triple. Another line of work that includes Minerva~\cite{MINERVA2018}, DeepPath~\cite{deeppath2017} explores reinforcement learning that trains an agent to find reasoning paths for KG completion. Another approach is to exploit attention over relations in a given KG  in order to learn end-to-end differentiable models~\cite{NeuralLP2017, Drum2019}. The final set of works include UNIKER~\cite{UNIKER2021}, RUGE~(\citeyear{RUGE2018}) that integrate traditional rules learnt via ILP models with KG embeddings models such that the two paradigms enhance each other's performance. Our approach is generic and exploits the rules learnt by the aforementioned models in order to enhance them further through proposed augmentations.
%We consider the setting of neuro-symbolic KB embeddings in our work where a combination of logical rules and knowledge base embeddings is utilized to reason about the validity of a triple. 
%Since our experiments are based on RNNLogic due to its high performance, we describe this in some detail.
%We consider RNNLogic~(\citeyear{RNNLogic2021}), a recent model, as our base model which is discussed below. 

\vspace{0.5ex}
 \noindent \textbf{RNNLogic+:} As a pre-processing step, for every $r\in \mathcal{R}$, RNNLogic adds a relation $r^{-1}$ to $\mathcal{R}$, and corresponding facts using inverse relations to $\mathcal{T}$. RNNLogic first produces a set of first order rules ($\mathcal{L}$) using an LSTM which are  
  used by the RNNLogic+ predictor to compute the score of a given triple. 
 Given a query $(\fol{h},\fol{r},\fol{?})$, the candidate answer $\fol{o}$ is scored by RNNLogic+ as:
\vspace{-0.08in}
\begin{equation}
\label{eq:RNNLogicplus score}
\fol{scor}(\fol{o}) = \fol{MLP}\big(\fol{PNA}(\{\textbf{v}_{\fol{l}} \,\vert \, \#(\fol{h}, \fol{r}, \fol{o})\}_{\fol{l} \in \mathcal{L} })\big)
\vspace{-0.08in}
\end{equation}
where the learnable embedding $\textbf{v}_{\fol{l}}$ of a given rule $\fol{l} \in \mathcal{L}$ is weighted by the number of groundings (\#) that triple ($\fol{h}, \fol{r}, \fol{o}$) satisfies in the rule $\fol{l}$'s body. The resulting weighted embeddings of all rules are aggregated by employing PNA aggregator~\cite{PNAAggregator2020} and this aggregated embedding is passed through an MLP to obtain a final score. 
%The parameters of the rule embeddings and MLP are trained by taking the softmax of score over all the candidate answers' scores. 

%The model's results can further be improved by introducing a KGE embedding model such as RotatE~\cite{sun2018rotate} into the scoring function resulting in the following score:
%Profiting from the success of KG embedding models,
The authors designed another scoring function that incorporates RotatE~\cite{sun2018rotate} into the scoring function, $\fol{scor}(\fol{o})$, in equation (\ref{eq:RNNLogicplus score}) where the goal is to exploit the knowledge encoded in the KG embeddings. The resulting scoring function is:
%\scalebox{0.9}{\parbox{.5\linewidth}{%  
\vspace{-0.08in}
\begin{equation}
%\begin{equation}
\label{eq: RNNlogicplus and KGE score}
\fol{score}_{\fol{KGE}}(\fol{o}) = \fol{scor}(\fol{o}) + \eta \, \fol{RotatE}\,(\fol{h}, \fol{r}, \fol{o})
%\end{equation}
%\vspace{-0.08in}
\end{equation}
where $\fol{RotatE}\,(\fol{h}, \fol{r}, \fol{o})$ is the score of the triple obtained from RotatE , and $\eta$ is a hyper-parameter. $\fol{RotatE}\,(\fol{h}, \fol{r}, \fol{o})$ is the negation of the value obtained by rotating the embedding for $\fol{h}$ by the rotation transformation defined by the embedding of $\fol{r}$ in complex space and computing the distance from the embedding of $\fol{t}$. Please refer to Appendix \ref{appendix: rotate} for further details.
%\paragraph{RLogic:}
%RLogic~(\citeyear{Rlogic2022}) is a recent rule generation model. In order to learn a rule: $\textbf{r}_{b}\Rightarrow \textbf{r}_{t}$, RLogic works in two steps: (i) first, it learns the embedding $\textbf{r}_{h}$ of ideal rule head, by recursively composing a relation in $\textbf{r}_{b}$ with the partial representation of body so far. (ii) ideal head representation $\textbf{r}_{h}$ and real head representation $\textbf{r}_{t}$ in KG are further passed to a neural network to measure the closeness between them. The top rules, whose $\textbf{r}_{h}$ and $\textbf{r}_{b}$ are close in step (i), are extracted after training.
%\paragraph{Related Work:} Recent work on Neuro-Symbolic KG Completion mainly comprise of models that exploit variational inference ~\cite{ExpressGNN2020}, reinforcement learning ~\cite{MINERVA2018}, attention-based end-to-end differentiable models ~\cite{NeuralLP2017} and models exploiting rules from traditional ILP models ~\cite{RUGE2018}.

\vspace{0.5ex}
\noindent \textbf{ExpressGNN:}
It is a novel model that integrates Markov Logic Networks (MLN) ~\cite{MLN2006} and Graph Neural Networks (GNN)~\cite{GNN2KipfandWelling017} to exploit their complementary strengths. An open-world paradigm is adopted in which a fact that is unknown in KG is assumed to be hidden (not false). The joint distribution of the observed and hidden triples of the KG in the MLN is optimized by employing a variational EM framework where the variational posterior distribution of the hidden variables is encoded as a GNN. Please refer to~\cite{ExpressGNN2020} for further details about the model.

\vspace{0.5ex}
 \noindent \textbf{PCA Score:} It is a symbolic rule confidence metric proposed in AMIE~(\citeyear{AMIE2013}) -- see Appendix \ref{appendix:PCAScoringFunction} for details. Broadly, it is the number of positive examples satisfied by a rule, divided by the total number of tails reached by the rule from heads occurring in the training dataset. Its performance in the context of AMIE was not as good due to its purely symbolic approach, and we are likely the first to show its utility in the context of NS-KGC.

%\subsection{Abduction}
\section{Rule Augmentation in NS-KGC Models}
\label{sec:main}
%The focus of our work is to propose two approaches to augment the rules already learnt by a given Neuro-Symbolic KG model in order to enhance the performance of a given model. Consequently, we explore how applying abduction and rule inversion on a given set of rules help in acquiring more rules without any overhead. We explain the two approaches in detail below. 
 With the aim of maximal utilization of a given rule $\fol{l} \in \mathcal{L}$, we first propose two rule augmentation techniques: abduction and rule inversion. The other two techniques prune low-quality rules from $\mathcal{L}$, and independently add new rules to increase coverage. All  augmentations are generic and can be integrated with any existing ruleset, and NS-KGC model.
 %that produce more rules from rules already generated by a given NS-KGC model.  
 
\vspace{0.5ex}
\noindent \textbf{Abduction:} The goal of abductive reasoning (or abduction) is to find the best explanation from a given set of observations~\cite{Pierce35}. It has seen limited use in the context of KBs \cite{yoshikawa-aaai19}. 
%Consequently, abduction helps in determining a set of \emph{assumptions} that should be made in order to deduce the required observations.
%For instance, in the above example, if we \emph{assume} $\fol{PlaceInCountry(Mississippi,US)}$ to be true in KG, the above rule would get fired given the other two observations.  
In our approach, for every rule in $\mathcal{L}$, we introduce several abductive rules with one of the antecedants, appearing as a consequent. As an example, 
%we explicitly encode an \emph{assumption} in what we call an \emph{abductive} rule. While designing an abductive rule, the inverse of assumed relation forms the head and the remaining relations in the original rule form the body.
%Traditionally, abductive reasoning has been based on first order logic rules. 
%For instance, 
consider the rule: 
\vspace{-0.08in}
\begin{equation*}
\fol{R1(X,Y) \wedge R2(Y,Z) \wedge R3(Z, W) \Rightarrow RH(X, W)}
\vspace{-0.08in}
\end{equation*}
Our augmentation will generate abductive rules, one for each relation in the body, as:
\vspace{-0.05in}
\begin{flalign*}
\fol{ R2(Y,Z) \wedge R3(Z,W) \wedge RH^{-1}(W,X)   \Rightarrow R1^{-1}(Y,X)} & \\
\fol{R3(Z, W) \wedge RH^{-1}(W,X) \wedge R1(X, Y) \Rightarrow R2^{-1}(Z,Y)} &\\
\fol{RH^{-1}(W, X) \wedge R1(X, Y) \wedge R2(Y, Z)   \Rightarrow R3^{-1}(W,Z)} 
\vspace{-0.08in}
\end{flalign*}

As an example, let's say a learned rule is $\fol{BornIn(X,U) \wedge PlaceInCountry(U,Y) \Rightarrow} \fol{Natio}$ $\fol{nality(X,Y)}$. If in the KG, we know that Oprah has nationality U.S., and that she is born in Mississippi, then abduction allows the model to hypothesize that Mississippi might be in U.S. Of course, not all abductions are accurate, for instance, just because Alabama is known to be in U.S., does not mean that Oprah was born in Alabama. Abductive rules increase rule coverage at the cost of precision. We expect the predictor scorer to automatically handle which (abductive) rules can and cannot be trusted.

\begin{table*}[t]
 \caption{Results of reasoning on four datasets with RNNLogic+ ($\fol{RNN}$). $\fol{Orig}$ represent RNNLogic rules. $\fol{RotE}$ represents RotatE. $\fol{AUG}$ represents our proposed augmentations. $\fol{RW}$ denotes rules discovered by random walks. }
 \vspace{-0.07in}
 \label{tab:tablemainMRRAndHitsAt10}
 \centering
\small
\begin{center}
%\begin{tabular}{*{11}{c|}} 
\begin{tabular}{|p{3.3cm}|p{0.6cm}p{0.5cm}p{0.65cm}|
p{0.6cm}p{0.5cm}p{0.65cm}|
p{0.6cm}p{0.5cm}p{0.65cm}|
p{0.6cm}p{0.5cm}p{0.65cm}|} 
%\Xhline{3\arrayrulewidth}
\Xhline{3\arrayrulewidth} 
\multirow{2}{2em}{\textbf{Algorithm}} & \multicolumn{3}{c|}{\textbf{WN18RR}} &
\multicolumn{3}{c|}{\textbf{FB15K-237}} &  \multicolumn{3}{c|}{\textbf{Kinship}} & \multicolumn{3}{c|}{\textbf{UMLS}}\\
%\cline{2-13}
 & MRR & H@1 & H@10 & MRR & H@1 &  H@10 & MRR & H@1 &  H@10 & MRR & H@1 & H@10 \\
%\Xhline{3\arrayrulewidth}
\Xhline{3\arrayrulewidth}
$\fol{[RNN]}$-$\fol{(RW)}$ & 44.2 & 41.6& 48.7 & 26.4 & 19.8&39.9 & 63.2 & 47.8 & 93.7 & 74.7 & 63.1 & 93.0  \\
%   \cline{1-13}
 $\fol{[RNN]}$-$\fol{(RW}$+$\fol{AUG)}$ & 47.7 & 44.3 & 54.3 & 29.5 & 21.5& 45.3 & 65.7 & 50.9 &94.8 & 79.7 & 69.5& 95.7 \\
   \cline{1-13}
%\Xhline{3\arrayrulewidth}
$\fol{[RNN}$+$\fol{RotE]}$-$\fol{(RW)}$ & 48.7 & 45.1&55.9 & 30.8 &22.8 & 46.9 & 71.4 & 58.0 & 95.7 & 82.0 & 73.5 & 95.3\\
%   \cline{1-13}
$\fol{[RNN}$+$\fol{RotE]}$-$\fol{(RW}$+$\fol{AUG)}$ & 51.1 & 47.4 &58.5 & 31.4 & 23.3&47.9 & 71.9 & 58.9& 96.2 & 83.8 & 75.8 &96.4\\
%\Xhline{3\arrayrulewidth}
\Xhline{3\arrayrulewidth}
 $\fol{[RNN]}$-$\fol{(Orig)}$ &49.6 & 45.5& 57.4 & 32.9 &24.0 & 50.6 & 61.6 &46.3 & 91.8 & 81.4 & 71.2 & 95.7 \\
 % $\fol{[RNN]}$-$\fol{(Orig}$+$\fol{AUGM)}$ & 51.9 & 47.6 & 60.3 & 33.8 & 25.1 & 51.4 & 66.9 & 52.5 & 94.1 & 82.6 & 74.1 & 96.4 \\
% \cline{1-13}
$\fol{[RNN]}$-$\fol{(Orig}$+$\fol{AUG)}$ & 52.7 & 48.3& 61.3 & 34.5 & 25.7& 51.9 & 68.7 & 54.8 & 95.7 & 84.0 & 75.2 & 96.4\\
 \cline{1-13}
%   \Xhline{3\arrayrulewidth}
$\fol{[RNN}$+$\fol{RotE]}$-$\fol{(Orig)}$ & 51.6 & 47.4& 60.2 &34.3  & 25.6& 52.4& 68.9 & 54.9 &94.6 & 81.5 & 71.2 &96.0 \\
% $\fol{RNN}$-$\fol{RotE}$-$\fol{AUGM}$ & 54.6 & 50.1 & 63.2 & 34.7 & 26.1 & 52.7 & 70.7 & 57.1 & 95.6 & 82.8 & 73.2 & 96.5\\
 %  \cline{1-13}
$\fol{[RNN}$+$\fol{RotE]}$-$\fol{(Orig}$+$\fol{AUG)}$ & \textbf{55.0} & \textbf{51.0} & \textbf{63.5}  & \textbf{35.3} & \textbf{26.5}&\textbf{52.9} & \textbf{72.9}& \textbf{59.9} & \textbf{96.4} & \textbf{84.2} & \textbf{76.1} & \textbf{96.5} \\
%\Xhline{3\arrayrulewidth}
\Xhline{3\arrayrulewidth}
\end{tabular}
\end{center}
\vspace{-1ex}
\end{table*}

\vspace{0.5ex}
\noindent \textbf{Rule Inversion:}
%As the rule generation is costly, our goal in this work is to harness maximal knowledge from a given superior rule. To this end, we propose our second approach to rule augmentation based on rule inversion. 
%With the focus on distilling more knowledge from a superior rule, 
Our second rule augmentation takes an existing rule and rewrites it by referring to inverses of all relations. 
%We now propose our second rule augmentation technique based on rule inversion. The key idea is that whenever a forward path between a pair of entities is true, we assume the backward path between them to be true. 
As an example, if a rule uses the path
$\fol{Oprah \xrightarrow{BornIn} Mississippi \xrightarrow{PlaceInCountry} US}$, then it could also use the equivalent path $\fol{US \xrightarrow{PlaceInCountry^{-1}} Mississippi \xrightarrow{BornIn^{-1}}}$ $\fol{Oprah}$. 
%Essentially, we regard each forward path to be bi-directional.
Formally, for every original rule:
\vspace{-0.05in}
\begin{equation*}
\fol{R1(X,Y) \wedge R2(Y,Z) \wedge R3(Z,W) \Rightarrow RH(X,W)}
\vspace{-0.05in}
\end{equation*}
we add to the ruleset the following inverted rule:
\vspace{-0.05in}
%\begin{flalign*}
%\fol{R3^{-1}(W,Z) \wedge R2^{-1}(Z,Y) \wedge R1^{-1}(Y,X) \Rightarrow RH^{-1}(W, X)}
%\end{flalign*}
\begin{equation*}
\resizebox{0.97\hsize}{!}{%
$\fol{R3^{-1}(W,Z) \wedge R2^{-1}(Z,Y) \wedge R1^{-1}(Y,X) \Rightarrow RH^{-1}(W, X)}$%      
}
\vspace{-0.05in}
\end{equation*}

\noindent \textbf{Rule Filtering:} 
Augmentations increase the size of the ruleset. 
In order to reduce the number of parameters and the training/test times of the NS-KGC model, we prune seemingly low-quality rules from the augmented rulebase. For this, we compute the PCA score for each original and augmented rule and prune all the rules that have score less than a threshold (set at 0.01 in experiments) and have less than 10 groundings. So, all low-coverage rules with seemingly low quality are pruned out. As experiments show, this results in up to 70\% reduction in the number of rules, while preserving KGC performance. 
% Essentially, it is the number of positive examples satisfied by the rule divided by the total number of $\fol{(x,y)}$ satisfied by the rule such that $\fol{r(x,y')}$ is a positive example for some $\fol{y'}$. 

\vspace{0.5ex}
\noindent \textbf{Random Walk Augmentation:} Motivated by the empirical success of PCA scores for finding good rules in the previous step, we further augment our ruleset with new, high scoring rules generated independently via local random walks. Starting at each entity in the KG, we perform a number of random walks of fixed length. Each such random walk constitutes the body of the rule and the relation connecting the end entities in the KG form the head of the discovered rule. We score these rules by the PCA score and retain all such rules that have PCA score above the threshold (of 0.1).

%For example, once RNNLogic model has trained and has generated the desired rules, our approach would first perform rule augmentation of each original rule by abduction and inversion. It would further filter out the best rules from all the rules. Finally, it would augment the rulebase by PCA-based random walk augmentation.
%Our approach to rule augmentation is generic and can be integrated with any of the existing NS-KG models. For example, once RNNLogic model has trained and has generated the desired rules, our approach would enhance these rules by designing abductive and inverted rules corresponding to each original rule.  It would further augment the rulebase by PCA-augmented rule generation. Finally, it would filter out the high-quality rules which would be passed on to RNNLogic+ predictor for training.
%\begin{enumerate}
% \item I think we would most likely draw a block diagram of the RNNLogic model and explain how abduction and inverse rules are incorporated into the model. Then we explain the block diagram by emphasizing on the abduction and inverse relations step

%\item Similarly, we would draw a block diagram of RLogic and explain how at what point abduction is introduced in the model.

%\item We could use a running example and explain abduction and inverse rules corresponding the running example
%\end{enumerate}
%\vspace{-0.1in}
\section{Experiments}
\label{sec:Experiments}
\noindent \textbf{Datasets:} We use four datasets for evaluation: WN18RR~\cite{WN18RRDataset2018}, FB15K-237~\cite{FB15K237dataset2015}, Kinship and UMLS~\cite{UMLSdataset2007}.  For each triple in test set, we answer queries $\fol{(h, r, ?)}$ and $\fol{(t, r^{-1}, ?)}$ with answers $\fol{t}$ and $\fol{h}$. We report the Mean Reciprocal Rank (MRR) and Hit@k (H@1, H@10) under the filtered measures~\cite{TransE2013}. Details and data stats are in Appendix \ref{appendix: datastatistics}.

%For each triplet $\fol{(h, r, t)}$ in the test set, traditionally queries of the form $\fol{(h, r, ?)}$ and $\fol{(?, r, t)}$ are created for evaluation, with answers $\fol{t}$ and $\fol{h}$ respectively. We model the $\fol{(?, r, t)}$ query as $\fol{(t, r^{-1}, ?)}$ with the same answer $\fol{h}$, where $\fol{r^{-1}}$ is the inverse relation for $\fol{r}$. In order to train the model over the inverse relations, we similarly augment the training data with an additional $\fol{(t, r^{-1}, h)}$ triple for every triple $\fol{(h, r, t)}$ present.

%Given ranks for all queries, we report the Mean Rank (MR), Mean Reciprocal Rank (MRR) and Hit@k (H@k, k = 1, 3, 10) under the filtered setting ~\cite{TransE2013}. To maintain consistency with RNNLogic, in cases where the model assigns same probability to other entities along with the answer, we compute the rank as $(m + \frac{(n+1)}{2})$ where $m$ is the number of entities with higher probabilities than the correct answer and $n$ is the number of entities with same probability as the answer.

%\paragraph{Experimental Setup:} 

\vspace{0.5ex}
\noindent \textbf{Baselines:} We first experiment with two base models: $\fol{RNNLogic+}$ ($\fol{[RNN]}$ in tables), and $\fol{RNNLogic+}$ with $\fol{RotatE}$ ($\fol{[RNN}$+$\fol{RotE}]$) (Eqn \ref{eq: RNNlogicplus and KGE score}). We have reproduced the numbers published by the original authors for these models (details in Appendix \ref{appendix: rnnlogicresultsreproduction}).
We run these models with two rulesets: (1) $\fol{Orig}$, rules generated by RNNLogic (around 300 rules per relation for WN18RR and FB15k-237, and 1000 rules per relation for Kinship and UMLS), and (2) $\fol{RW}$, only the rules discovered by our random walks. This second setting can only evaluate the value of abduction, inversion, and pruning since random walks are anyways used in generating rules.
%which is further trained with $\fol{RNNLogic+}$ ($\fol{RNN}$) (\ref{eq:RNNLogicplus score}) and $\fol{RNNLogic+}$ with $\fol{RotatE}$ ($\fol{RNN}$-$\fol{RotE}$) (\ref{eq: RNNlogicplus and KGE score}). To test the utility of abduction and rule inversion with a different set of rules, we utilize the rules discovered by random walk augmentation with RNNLogic+, both with ($\fol{RW}$-$\fol{RotE}$) and without ($\fol{RW}$) $\fol{RotatE}$.  
 More details in Appendix \ref{sec:rulestatistics}, \ref{appendix:rulegenerationviarandomwalks}  and \ref{appendix:RNNLogic+training}. 

In order to assess the generality of our augmentations, we also experiment with ExpressGNN ~\cite{ExpressGNN2020}. We choose top five rules for each relation from RNNLogic's $\fol{Orig}$ ruleset according to PCA confidence and provide them as input ruleset to ExpressGNN ($\fol{[ExpGNN]}$ in tables). ExpressGNN does not scale up to the augmented ruleset for FB15K-237, hence we test it for the other three datasets. 
Refer to Appendix \ref{appendix: expressGNNbaseline} for more details. We use $\fol{AUG}$ to denote the performance of rule augmentations for all baselines.

We also tried rulesets from NeuralLP (\citeyear{NeuralLP2017}), but they are too small to be useful with RNNLogic+. 
The only other NS-KGC model that has reported performance similar to RNNLogic+ is RLogic~(\citeyear{Rlogic2022}). Unfortunately, their code is not publicly available.\footnote{Our reimplementation could not match reported results, and sending several emails to original authors was not helpful.}

\vspace{0.5ex}
\noindent \textbf{Results: } We report the results in Table \ref{tab:tablemainMRRAndHitsAt10} for the RNNLogic baselines (further details in Appendix \ref{appendix:RNNLogic+results}). We observe that in all settings, there is a notable increase in performance using augmented rules. In particular, we obtain 7.1 pt and 8.5 pt increase in MRR and Hits@1 in  $\fol{[RNN]}$-$\fol{(Orig)}$ setting on Kinship, and 3.5 pt and 5.6 pt increase in MRR and Hits@10 in $\fol{[RNN]}$-$\fol{(RW)}$ setting for WN18RR dataset. 
 %The groundings from the new rules can be used to create new paths between entities in KG, resulting in more informed scores as computed RNNLogic+. 
We also find that rule augmentations complement RotatE scores in capturing more information about the KG, leading to improved performance in those settings too. To the best of our knowledge, our best results for WN18RR are state-of-the-art for NS-KGC models.%for a given model (as seen in $\fol{[RNN}$+$\fol{RotE]}$ and $\fol{[RNN}$+$\fol{RotE]}$-$\fol{(RW)}$).
%We see that in both RNNLogic+ (row 1 vs 2) and RNNLogic+ with RotatE (row 3 vs 4), there is a significant increase in performance across datasets after performing rule augmentation. In particular, we obtain 23\% improvement in the MR over RNNLogic+ on UMLS and 3.4 points of improvement in MRR over RNNLogic+ on WN18RR. We claim that the reason for this is that our rule augmentation techniques result in better utilization of the learned high-quality rules from RNNLogic. The groundings from the new rules can be used to create new paths between entities in the knowledge graph, resulting in more informed scores as computed by RNNLogic+. Further, rule augmentation also complements RotatE ensembling (row 4) in capturing more information about the KG for the model. 

Next, we present the results of our proposed augmentations with ExpressGNN\footnote{\url{https://github.com/expressGNN/ExpressGNN}} as baseline in Table \ref{tab:expressGNNresults}. We note that ExpressGNN assumes the knowledge of test queries while it constructs the MLN during training. Therefore, the results presented in Table \ref{tab:expressGNNresults} are not directly comparable with the results of other models presented in the paper, which do not make this assumption. We observe substantial gains on all datasets and all metrics, notably a 22.4 pt MRR, 17.9 pt Hits@1 and 29.9 pt Hits@10 improvement on WN18RR dataset with our augmentations ($\fol{AUG}$). This experiment demonstrates that $\fol{AUG}$ can help other neuro-symbolic settings as well. Refer to Appendix \ref{appendix: expressGNNbaseline} for more details.
\begin{table}[h]
%\begin{threeparttable}[b]
\caption{Results of reasoning on three datasets with ExpressGNN ($\fol{ExpGNN}$). $\fol{AUG}$ represents our proposed augmentations%\tnote{1}
\footnotemark. 
%\tablefootnote{Please note that results in this table are not directly comparable to results in Table \ref{tab:tablemainMRRAndHitsAt10}}
}
\centering
\small
\begin{center}
\begin{tabular}{|p{1.2cm} p{1.7cm}p{0.65cm}p{0.65cm}p{0.70cm}|} 
 \cline{1-5}
\textbf{Dataset} & \textbf{Model} & \textbf{MRR} & \textbf{H@1} & \textbf{H@10} \\
 \cline{1-5}
\multirow{2}{4em}{WN18RR} & $\fol{[ExpGNN]}$	& 52.3 & 44.1 &	63.6\\
& $\fol{[ExpGNN}$+$\fol{AUG]}$	& \textbf{74.7} & \textbf{62.0} &  \textbf{93.5} \\
 \cline{1-5}
\multirow{2}{4em}{UMLS} & $\fol{[ExpGNN]}$ &	58.1 & 44.4 &	77.6 \\
& $\fol{[ExpGNN}$+$\fol{AUG]}$ &	\textbf{60.9} &	\textbf{49.2} & \textbf{83.4} \\
\cline{1-5}
\multirow{2}{4em}{Kinship} & $\fol{[ExpGNN]}$ &	52.7 & 41.7 &	79.8 \\
& $\fol{[ExpGNN}$+$\fol{AUG]}$ &	\textbf{64.1} & \textbf{49.5} & \textbf{93.2} \\
\cline{1-5}
\end{tabular}
%\begin{tablenotes}
%       \item [1] Please note that results in this table are not directly comparable to results in Table \ref{tab:tablemainMRRAndHitsAt10}
%\end{tablenotes}
\end{center}
%\end{threeparttable}
\vspace{-0.20in}
\label{tab:expressGNNresults}
\end{table}
\footnotetext{ Please note that results in this table are not directly comparable to results in Table \ref{tab:tablemainMRRAndHitsAt10}}
% \vspace{-0.10in}
\section{Analysis of Augmented Rules}
\label{sec:analysis of results}
We perform five further analyses to answer the following questions. \textbf{Q1.} Are the rules created by abduction and rule inversion of high quality? \textbf{Q2}. What is the individual effect of each type of augmentation on the performance? \textbf{Q3}. How do the rule augmentations affect the training time of a model? \textbf{Q4}. Can we get the same performance as augmentation by generating more rules from the LSTM in RNNLogic? \textbf{Q5}. Are the augmented rules interpretable by a human?

\vspace{0.5ex}
\noindent \textbf{Quality of New Rules:} To answer \textbf{Q1}, we employ two metrics to assess quality of rules, (PCA-metric and FOIL-metric) before and after abduction and rule inversion. The rules obtained from random walks have high scores by construction since they are filtered based on PCA score. Therefore, they are of high quality as per our definition. (Details in Appendix \ref{appendix:PCAScoringFunction} and \ref{appendix:FOILScoringFunction}) 

\vspace{-0.10in}
\begin{table}[H]
 \caption{Number of high quality rules before and after augmentations on rules generated by RNNLogic.}
  \vspace{-0.07in}
 \label{tab:tableFOILandPCA}
 \centering
\small
\begin{center}
%\begin{tabular}{*{11}{c|}} 
\scalebox{0.88}{
\begin{tabular}{|c|cc|cc|} 
\Xhline{3\arrayrulewidth}
%\Xhline{3\arrayrulewidth} 
\multirow{2}{4em}{\textbf{Rule Set}} & 
\multicolumn{2}{c|}{\textbf{WN18RR}} &
\multicolumn{2}{c|}{\textbf{UMLS}} \\
%\cline{2-5}
& FOIL & PCA & FOIL & PCA \\
\Xhline{3\arrayrulewidth}
%\Xhline{3\arrayrulewidth}
$\fol{Original}$ & 2286 & 2647 & 25079 & 28982 \\
%\cline{1-5}
$\fol{Original}$ w/ $\fol{INV}$ & 3157 & 3577 & 42188 & 46908 \\
%\cline{1-5}
$\fol{Original}$ w/ $\fol{ABD}$ & 7141 & 7607 & 68693 & 84554 \\
%\cline{1-5}
$\fol{Original}$ w/ $\fol{INV}$ + $\fol{ABD}$ & \textbf{8502} & \textbf{9155} & \textbf{100146} & \textbf{125019} \\
\Xhline{3\arrayrulewidth}
%\Xhline{3\arrayrulewidth}
\end{tabular}
}
\end{center}
\vspace{-0.15in}
\end{table}

Table \ref{tab:tableFOILandPCA} presents the number of rules that have a score of at least 0.1 according to each metric, which we regard as criteria for defining a high-quality rule. We observe that there is a large increase in the number of high-quality rules after abduction and rule inversion, nearly tripling in the case of abduction (row 1 vs row 3).  This is because the augmented rules exploit the same groundings as the original rules, in the form of new rules. Thus, augmented counterparts of high-quality rules are likely to be high-quality. Overall, we find that abduction and rule inversion does indeed produce high-quality rules. 
%These groundings can be used in the context of different relations in the rule head, after abduction and inversion. 
%Interestingly, we find that the ratio of high-quality rules to the total number of rules mostly goes down after the rule augmentation. However, the model still gives better performance. This is because the RNNLogic+ score considers only the number of groundings of the rule. If this is nearly 0 for all possible tails given a head, then it will not affect the model. Our rule scoring function assigns low scores to all such rules. Thus, adding rules that have low scores according to our metrics does not affect performance for RNNLogic+. We confirmed this by running the model on the filtered rule set which is obtained by considering only the rules whose score is above $0.1$. In all cases, we obtained performance within 1 point MRR of the full rule set. This opens up opportunities for parameter efficiency as the number of rule embeddings required by the model goes down by 50-70\% when we consider only the filtered rules. 

%It also indicates that the scores defined by us are a good indicator of the utility of a rule. 

\vspace{0.5ex}
\noindent \textbf{Ablation:} To answer \textbf{Q2}, we perform an ablation study for inversion ($\fol{INV}$), abduction ($\fol{ABD}$), random walk augmentation ($\fol{RW}$) and rule filtering ($\fol{FIL}$) on $\fol{[RNN}$+$\fol{RotE]}$-$\fol{(Orig)}$ setting for WN18RR and Kinship datasets to observe the impact of each type of augmentation. The results are presented in Table \ref{tab:tableAblationStudy} (further details are in Appendix \ref{appendix:RNNLogic+ablation}).

In general, abduction (row 3) gives larger improvements than rule inversion (row 2) because as we noticed in the previous section, abduction adds a larger number of high-quality rules to the rule set. We also find that adding the PCA-based random walk rules results in performance improvement, even with only 5\% new rules being added (as in Kinship) as compared to original rule set. Finally, we find that filtering based on the PCA metric results in marginal performance improvement, along with lower running times (see below). 
%(\textbf{Q3})

%\vspace{-0.04in}
\begin{table}
  \caption{Ablation study on WN18RR and Kinship for filtering ($\fol{FIL}$), inversion ($\fol{INV}$), abduction ($\fol{ABD}$) and PCA-filtered random walk augmentation ($\fol{RW}$). 
  %$\fol{AUG}$ represent all the four augmentation techniques performed on RNNLogic+ with RotatE model.
  }
\vspace{-0.07in}
 \label{tab:tableAblationStudy}
 \centering
\small
\begin{center}
\scalebox{0.88}{
%\begin{tabular}{*{11}{c|}} 
\begin{tabular}{|p{1.9cm}|p{0.56cm}p{0.5cm}p{0.65cm}|
                          p{0.56cm}p{0.5cm}p{0.65cm}|} 
%\Xhline{3\arrayrulewidth}
\Xhline{3\arrayrulewidth} 
\multirow{2}{2em}{\textbf{Algorithm}} & 
\multicolumn{3}{c|}{\textbf{WN18RR}} &
\multicolumn{3}{c|}{\textbf{Kinship}} \\
%\cline{2-7}
& MRR & H@1 & H@10 & MRR & H@1 & H@10  \\
\Xhline{3\arrayrulewidth}
%\Xhline{3\arrayrulewidth}
$\fol{AUG }$ & \textbf{55.0} & \textbf{51.0} & \textbf{63.5} & \textbf{72.9} & \textbf{59.9} & \textbf{96.4} \\
 %  \cline{1-7}
 $\fol{ AUG\, \, minus \, \, ABD}$ & 52.2 & 47.8 & 61.0 & 71.3 & 57.8 & 96.2 \\
%\cline{1-7}
$\fol{ AUG\, \, minus \, \, INV}$ & 54.4 & 50.0 & 62.7 & 71.3 & 57.7 & \textbf{96.4} \\
%\cline{1-7}
$\fol{ AUG\, \, minus \, \, FIL}$ & \textbf{55.0}& 50.6 & 63.3& 72.5 & 59.5 & \textbf{96.4}  \\
%\cline{1-7}
$\fol{ AUG\, \, minus \, \, RW}$ & 54.6 & 50.1 & 63.2 & 70.7 & 57.1 & 95.6 \\ 
\Xhline{3\arrayrulewidth}
%\Xhline{3\arrayrulewidth}
\end{tabular}
}
\end{center}
\vspace{-0.10in}
\end{table}
\begin{table}
 \caption{Table showing performance/time trade-off per epoch on two datasets. T/T(min) represents training time per epoch in minutes.}
 \label{tab:performancetimetradeoff}
\centering
\small
\begin{center}
\resizebox{\columnwidth}{!}{
\begin{tabular}{|p{1.0cm}p{2.3cm} p{0.8cm}p{0.5cm}p{0.6cm}p{0.5cm}p{0.75cm}|} 
\cline{1-7}
\textbf{Dataset} &	\textbf{Modification} &	\textbf{\#Rules} &	\textbf{T/T} &	\textbf{MRR} & \textbf{H@1} &	\textbf{H@10} \\
\cline{1-7}
\multirow{3}{4em}{WN18RR} &	$\fol{Orig}$ &	6135 &	334	& 51.6 &	47.4 &	60.2\\
& $\fol{Orig + AUG}$ &	25729 &	1520 &	55.0 &	50.6 &	63.3\\
& $\fol{Orig + AUG + FIL}$ & 20053 & 931 &	\textbf{55.0} &	\textbf{51.0} &	\textbf{63.5} \\
\cline{1-7}
\multirow{3}{4em}{Kinship} & $\fol{Orig}$ &49994 &	5 & 68.9 &	54.9 & 94.6\\
& $\fol{Orig + AUG}$ & 315865 & 36 & 72.5 & 59.5 & 96.4 \\
& $\fol{Orig + AUG + FIL}$ & 97331 & 11 & \textbf{72.9} & \textbf{59.9} & \textbf{96.4} \\
\cline{1-7}
\end{tabular}
}
\end{center}
\vspace{-0.10in}
\end{table}

\noindent
 \textbf{Performance vs Training Time Trade-off:} 
To answer \textbf{Q3}, we report training time per epoch (in minutes), size of ruleset and performance metrics after augmentation through $\fol{ABD}$, $\fol{INV}$ and $\fol{RW}$ (denoted as $\fol{AUG}$) and filtering ($\fol{AUG + FIL}$) with $\fol{[RNN + RotE]}$ as the baseline model in Table \ref{tab:performancetimetradeoff}.

Our proposed augmentations ($\fol{INV}$, $\fol{ABD}$ and $\fol{RW}$) result in substantial performance gains, at the cost of 5-6 times increase in the training time. After filtering ($\fol{FIL}$), there is no decrease in performance, and the training time goes down by 2-3$\times$ compared to $\fol{AUG}$. Therefore, we obtain substantial performance gains through our augmentations, at the cost of only 2-3 times increase in training time.

\vspace{0.5ex}
\noindent \textbf{Rule Generation vs Rule Augmentation:} 
Our augmentations result in 100-200\% increase in the number of rules across datasets after filtering. As a control experiment to answer \textbf{Q4}, we train RNNLogic to generate $80$ rules per relation (R/R) and augment resulting rules without filtering (for a fair comparison). We further train RNNLogic with $500$ rules per relation without augmentation and compare performance of both rulesets (which now have comparable size) using $\fol{[RNN}$+$\fol{RotE]}$ on WN18RR and Kinship in Table  \ref{tab:tableAugumentationvsgeneration} (see Appendix \ref{appendix:RuleAugmentationvsRuleGeneration}). %\todo{M: read this. }
%Although the number of abductive rules per relation vary, the average number of abductive rules per relation are $168$, $364$, $494$ for $100$, $200$, $300$ original relations respectively. 
% The hyperparameters of this experiment were optimized on validation set.
% %Note that the rule file generated by RNNLogic with 500 rules per relation is larger than the rule set with 100 rules per relation even with ABD. 
%\vspace{-0.06in}
\begin{table}
 \caption{Performance of augmentation on WN18RR and Kinship. R/R and TR is number of rules per relation and total rules generated from RNNLogic respectively.}
  \vspace{-0.07in}
 \label{tab:tableAugumentationvsgeneration}
 \centering
\small
\begin{center}
%\begin{tabular}{*{11}{c|}} 
\scalebox{0.88}{
\begin{tabular}{|p{1.10cm}|cccccc|} 
\Xhline{3\arrayrulewidth}
%\Xhline{3\arrayrulewidth} 
Dataset & R/R& TR & AUG & MRR & H@1 & H@10 \\
\Xhline{3\arrayrulewidth}
%\Xhline{3\arrayrulewidth}
\multirow{2}{4em}{WN18RR} & 80 & 9867 & Yes & \textbf{49.0}  & \textbf{44.9}  & \textbf{56.7}  \\
%\cline{2-7}
& 500 & 11000 & No & 47.7 & 43.7  & 55.2  \\
 \cline{1-7}
\multirow{2}{4em}{Kinship} & 80 & 18432 & Yes & \textbf{69.5} & \textbf{56.1} & \textbf{94.6}  \\
%\cline{2-7}
 & 500 & 25000 & No & 66.1 &52.1 & 93.1 \\
\Xhline{3\arrayrulewidth}
%\Xhline{3\arrayrulewidth}
\end{tabular}
}
\end{center}
\vspace{-0.15in}
\end{table}

We observe that rule augmentations lead to large improvement over rule generation in all cases, even when rule generation creates more rules. Thus, we find that rule augmentation is more beneficial than simply using more rules from the rule generator.
%This is because the rule weights and H(rule) metric used by RNNLogic to internally evaluate rules have a significant value for very few rules. On average, only around 35 rules per relation have a rule weight above 0.1 for WN18RR. 
Augmentations exploit a small number of high-quality rules to their full potential. 
%This opens the avenue for future work improving the rule evaluation metric in RNNLogic, thus identifying a better set of seed rules for augmentation.
%\vspace{0.5ex}

\vspace{0.5ex} 
\noindent \textbf{Qualitative Analysis:} To answer \textbf{Q5}, we randomly sample 50 rules from the $\fol{Orig}$ and $\fol{RW}$ rules for
the FB15K-237 dataset and score them as 0 (gibberish), 1 (logically dubious but statistically plausible) and 2 (logically correct) for each ruleset. The reported numbers are averages of scores obtained from two human annotators. We do not include $\fol{INV}$ and $\fol{ABD}$ in this comparison as they are generated from $\fol{Orig}$ rules utilizing the same groundings and thus we expect them to be as interpretable. The scores are 0.90 ($\fol{Orig}$) and 1.23 ($\fol{RW}$). $\fol{RW}$ rules are more interpretable due to their high PCA scores. One example of an interpretable rule added by $\fol{RW}$ is $\fol{Friends(A, \, C), Inverse\_Producer(C, \,D),}$ $ \fol{Writer(D}$ $\fol{,\, B)}$ :- $\fol{Friends(A, \, B)}.$ We provide additional rule examples for each type of augmentation in Appendix \ref{appendix:qualitativeanalysisofrules}. 

\section{Conclusion and Future Work}
\label{sec:conclusionandfuturework}
We present simple rule augmentation techniques in the context of Neuro-Symbolic Knowledge Graph models and obtain substantial increase in performance over strong base models. We believe our augmentations can become standard for all subsequent NS-KGC models. %We make effective use of existing rule scoring functions such as PCA for pruning of our augmented rules. and show their utility in the context of NS-KGC. 
%Rule augmentation makes better use of high-quality rules generated by rule generators such as RNNLogic. Our work opens several avenues for future work. 
We release code and rulesets for further research. Future work includes using our augmentation technique during the iterative learning of rules in algorithms such as RNNLogic, potentially further improving their performance.

 \section*{Limitations}
 \label{sec:limitations}
% ACL 2023 requires all submissions to have a section titled ``Limitations'', for discussing the limitations of the paper as a complement to the discussion of strengths in the main text. This section should occur after the conclusion, but before the references. It will not count towards the page limit.
% The discussion of limitations is mandatory. Papers without a limitation section will be desk-rejected without review.

% While we are open to different types of limitations, just mentioning that a set of results have been shown for English only probably does not reflect what we expect. 
% Mentioning that the method works mostly for languages with limited morphology, like English, is a much better alternative.
% In addition, limitations such as low scalability to long text, the requirement of large GPU resources, or other things that inspire crucial further investigation are welcome.
Since rule abduction and inversion utilize the same groundings as the original rules, Neuro-Symbolic KGC models that are based on grounding the entire rule will not benefit from these augmentations. Abduction and inversion also require the model to be trained on a knowledge graph that contains the inverse relations $\fol{r^{-1}}$ for each relation $\fol{r}$. Finally, since RNNLogic+ has a separate rule embedding for each rule, performing rule augmentation increases the number of parameters in the model and leads to longer training times and larger GPU memory consumption.

 \section*{Ethics Statement}
 \label{sec:ethicsstatement}
 We anticipate no substantial ethical issues arising due to our work on rule augmentation for Neuro-Symbolic KGC. Our work relies on a set of rules generated from another source to perform augmentation. This may result in the augmented rule set exaggerating the effect of malicious or biased rules in the original rule set. 
% Scientific work published at ACL 2023 must comply with the ACL Ethics Policy.\footnote{\url{https://www.aclweb.org/portal/content/acl-code-ethics}} We encourage all authors to include an explicit ethics statement on the broader impact of the work, or other ethical considerations after the conclusion but before the references. The ethics statement will not count toward the page limit (8 pages for long, 4 pages for short papers).

\section*{Acknowledgements}
\label{sec:acknowledgements}
This work is supported by IBM AI Horizons Network grant, grants by Google, Verisk, and 1MG, an IBM SUR award, and the Jai Gupta chair fellowship by IIT Delhi. We acknowledge travel support by Google and Yardi School of AI travel grants. We thank the IIT Delhi HPC facility for its computational resources.

\bibliography{anthology,custom}
\bibliographystyle{acl_natbib}

\appendix

\begin{table*} % MAIN TABLE BEGINS
% FIRST TABLE BEGINS
\centering
\small
\caption{Statistics of Knowledge Graph datasets}
\vspace{0.03in}
\begin{tabular}{|cccccc|}
\Xhline{3\arrayrulewidth}
%\Xhline{3\arrayrulewidth}
Datasets & \#Entities & \#Relations & \#Training & \#Validation & \#Test \\
\hline
FB15K-237 & 14541 & 237 & 272,115 & 17,535 & 20,446 \\
%\hline
WN18RR & 40,943 & 11 & 86,835 & 3,034 & 3,134 \\
%\hline
Kinship & 104 & 25 & 3,206 & 2,137 & 5,343 \\
%\hline
UMLS & 135 & 46 & 1,959 & 1,306 & 3,264 \\
%\Xhline{3\arrayrulewidth}
\Xhline{3\arrayrulewidth}
\end{tabular}
\label{tab:KG statistics}
\vspace{0.12in}
% FIRST TABLE ENDS

% SECOND TABLE STARTS
\centering
\small
\caption{RNNLogic rules used per dataset.  $\fol{INV}$ and $\fol{ABD}$, $\fol{RW}$ represent rule inversion and abduction and PCA-based walk rule augmentation respectively. The last column represents the rule filtering ($\fol{FIL}$) applied on all the rules.}
\vspace{0.03in}
\begin{tabular}{|cp{0.75cm}cp{0.95cm}ccc|}
%\Xhline{3\arrayrulewidth}
\Xhline{3\arrayrulewidth}
\multirow{2}{4em}{Datasets} & \multirow{2}{4em}{\#Rules}  &\#$\fol{Rules}$ & \#$\fol{Rules}$ & \#$\fol{Rules}$ + & \#$\fol{Rules}$ +$\fol{INV}$ & \#$\fol{Rules}$ +$\fol{INV}$+\\
 & & + $\fol{INV}$ & + $\fol{ABD}$ & $\fol{INV}$ + $\fol{ABD}$ &  + $\fol{ABD}$ + $\fol{RW}$ & $\fol{ABD}$ + $\fol{RW}$ + $\fol{FIL}$  \\ 
\Xhline{3\arrayrulewidth}
{FB15K-237} & {126137}  & {174658} & {295403} & {392280} & {394967} & {298446} \\
%\hline
{WN18RR} & {6135}  & {8742} & {18251} & {23304} &  {25729}  &  {20053} \\
%\hline
{Kinship}   &  {49994}    &  {91544}    &   {171302}    &   {301646}      &  {315865} &  {97331} \\ 
%\hline
{UMLS} & {91908}  & {171526} & {322464} & {564374} & {574687} & {204504} \\
\Xhline{3\arrayrulewidth}
%\Xhline{3\arrayrulewidth}
\end{tabular}
\label{tab:rule size}
\vspace{0.12in}
% SECOND TABLE ENDS

% THIRD TABLE STARTS
\centering
\small
\caption{Results of reasoning on four datasets: WN18RR, FB15K-237, Kinship and UMLS with RNNLogic+ ($\fol{RNN}$). $\fol{Orig}$ represents rules acquired from RNNLogic.  $\fol{RotE}$ represents RotatE. $\fol{AUG}$ represents all the proposed approaches in our work. $\fol{RW}$ represents rules obtained only from PCA-filtered random walk augmentation. }
\vspace{0.03in}
\begin{tabular}{|p{3.5cm}|ccccc|ccccc|} 
%\Xhline{3\arrayrulewidth}
\Xhline{3\arrayrulewidth} 
\multirow{2}{4em}{\textbf{Algorithm}} & \multicolumn{5}{c|}{\textbf{WN18RR}} &
\multicolumn{5}{c|}{\textbf{FB15K-237}} \\
%\cline{2-11}
& MR & MRR & H@1 & H@3 & H@10 & MR & MRR & H@1 & H@3 & H@10 \\
\Xhline{3\arrayrulewidth}
%\Xhline{3\arrayrulewidth}
$\fol{[RNN]}$-$\fol{(RW)}$ & 8218.73   & 44.2 &41.6  & 45.5& 48.7& 808.32 &  26.4& 19.8 &  28.9&  39.9\\
%\cline{1-11}
 $\fol{[RNN]}$-$\fol{(RW}$+$\fol{AUG)}$ & 7241.14 & 47.7  & 44.3 &49.2  & 54.3  & 481.58 &  29.5& 21.5 & 32.3 & 45.3\\
%\Xhline{3\arrayrulewidth}
$\fol{[RNN}$+$\fol{RotE]}$-$\fol{(RW)}$ & 4679.70   & 48.7  & 45.1 & 49.8 &55.9 & 521.06 & 30.8 &22.8 &33.5  & 46.9 \\
%\cline{1-11}
$\fol{[RNN}$+$\fol{RotE]}$-$\fol{(RW}$+$\fol{AUG)}$ & 4431.75  & 51.1 & 47.4  & 52.6  & 58.5 & 279.65 &   31.4  & 23.3 & 34.3 & 47.9\\ 
\Xhline{3\arrayrulewidth}
 $\fol{[RNN]}$-$\fol{(Orig)}$ & 5857.65 & 49.6 & 45.5& 51.4& 57.4& 256.14& 32.9&24.0 &36.1 & 50.6\\
%\cline{1-11}
% $\fol{RNN}$-$\fol{AUGM}$ & 5265.80 & 51.9  & 47.7 & 53.7 & 60.3 & 230.71 & 33.8 & 25.1 & 37.3 & 51.4\\
$\fol{[RNN]}$-$\fol{(Orig}$+$\fol{AUG)}$ & 5156.38 &52.7 &48.3  &54.9 &61.3 &218.11 &34.5 &25.7 &37.9  &51.9 \\
%\Xhline{3\arrayrulewidth}
$\fol{[RNN}$+$\fol{RotE]}$-$\fol{(Orig)}$ & 4445.79  &51.6  &47.4  &53.4 & 60.2&217.30 & 34.3 & 25.6  &37.5  & 52.4 \\
%\cline{1-11}
% $\fol{RNN}$-$\fol{RotE}$-$\fol{AUGM}$ &  4263.43 & 54.6 & 50.1  &  57.0  & 63.2 & 206.54 &  34.7  & 26.1 & 38.3 & 52.7\\ 
$\fol{[RNN}$+$\fol{RotE]}$-$\fol{(Orig}$+$\fol{AUG)}$  &\textbf{4231.77}  &\textbf{55.0} & \textbf{51.0} &\textbf{57.2}  &\textbf{63.5}  & \textbf{198.81} &\textbf{35.3}  & \textbf{26.5} &\textbf{38.7}  & \textbf{52.9}\\
\Xhline{3\arrayrulewidth}
\multirow{2}{4em}{\textbf{Algorithm}} & \multicolumn{5}{c|}{\textbf{Kinship}} & \multicolumn{5}{c|}{\textbf{UMLS}} \\
%\cline{2-11}
& MR & MRR & H@1 & H@3 & H@10 & MR & MRR & H@1 & H@3 & H@10 \\
\Xhline{3\arrayrulewidth}
%\Xhline{3\arrayrulewidth}
$\fol{[RNN]}$-$\fol{(RW)}$ & 3.6 & 63.2 & 47.8 &73.5 &93.7 & 5.17&  74.7& 63.1 &83.6 & 93.0  \\
%\cline{1-11}
 $\fol{[RNN]}$-$\fol{(RW}$+$\fol{AUG)}$  & 3.36 & 65.7 &50.9  &75.8  & 94.8 & 3.65 & 79.7  &69.5  & 87.8 & 95.7 \\
%\Xhline{3\arrayrulewidth}
$\fol{[RNN}$+$\fol{RotE]}$-$\fol{(RW)}$ & 2.99  &71.4  & 58.0 &81.6  &95.7 & 3.46 & 82.0 & 73.5 & 88.9  & 95.3 \\
%\cline{1-11}
$\fol{[RNN}$+$\fol{RotE]}$-$\fol{(RW}$+$\fol{AUG)}$ &2.89  & 71.9  & 58.9 & 81.7 & 96.2 & 3.20 & 83.8    & 75.8  & 90.0  & 96.4 \\ 
\Xhline{3\arrayrulewidth}
 $\fol{[RNN]}$-$\fol{(Orig)}$ &4.45  & 61.6 &46.3 &71.7 &91.8 &3.66 &81.4 & 71.2&90.3 &95.7 \\
%\cline{1-11}
% $\fol{RNN}$-$\fol{AUGM}$ & 3.58 &  66.9 & 52.5 & 77.00  & 94.1  & 3.18 & 82.6 & 74.1 & 91.0 & 96.4 \\
$\fol{[RNN]}$-$\fol{(Orig}$+$\fol{AUG)}$ & 3.15& 68.7& 54.8 &78.9 
 &95.7 & \textbf{2.81} &84.0 &75.2  & \textbf{91.5} & 96.4 \\
%\Xhline{3\arrayrulewidth}
$\fol{[RNN}$+$\fol{RotE]}$-$\fol{(Orig)}$ & 3.28  &68.9  & 54.9 &78.8 &94.6 &3.17 &81.5  &71.2  & 90.1 &96.0  \\
%\cline{1-11}
% $\fol{RNN}$-$\fol{RotE}$-$\fol{AUGM}$ & 2.99 & 70.7 & 57.1  & 80.8  & 95.6 & 3.05 & 82.8 & 73.2 & 91.1 & 96.5\\ 
$\fol{[RNN}$+$\fol{RotE]}$-$\fol{(Orig}$+$\fol{AUG)}$  & \textbf{2.80} &\textbf{72.9}  &\textbf{59.9}  &\textbf{82.6}  &\textbf{96.4}  & 2.83  &\textbf{84.2}  &\textbf{76.1}  &91.3  &\textbf{96.5} \\
\Xhline{3\arrayrulewidth}
\end{tabular}
\label{tab:appendixMainDetailedresults}
\vspace{0.12in}
% THIRD TABLE ENDS

% FOURTH TABLE STARTS
\centering
\small
\caption{Ablation study performed on Kinship and UMLS for filtering ($\fol{FIL}$), inversion ($\fol{INV}$), abduction ($\fol{ABD}$) and random walk augmentation ($\fol{RW}$). $\fol{AUG}$ represents all proposed approaches in our work taken together.}
\vspace{0.03in}
\begin{tabular}{|p{2.2cm}|ccccc|ccccc|} 
\Xhline{3\arrayrulewidth}
%\Xhline{3\arrayrulewidth}
\multirow{2}{4em}{\textbf{Algorithm}} & \multicolumn{5}{c|}{\textbf{Kinship}} & \multicolumn{5}{c|}{\textbf{UMLS}} \\
%\cline{2-11}
& MR & MRR & H@1 & H@3 & H@10 & MR & MRR & H@1 & H@3 & H@10  \\
\Xhline{3\arrayrulewidth}
%\Xhline{3\arrayrulewidth}
   $\fol{AUG }$ & \textbf{2.80}  & \textbf{72.9}  & \textbf{59.9}  & \textbf{82.6} & \textbf{96.4} & \textbf{2.83} & \textbf{84.2} & \textbf{76.1} & \textbf{91.3} & \textbf{96.5}  \\
%   \cline{1-11}
   $\fol{ AUG \, \,  minus \, \,  ABD }$ & 2.90 & 71.3 & 57.8 & 81.4 & 96.2 & 3.16 & 82.6 & 72.9  & 90.8 & 96.5 \\
%   \cline{1-11}
   $\fol{AUG \, \,  minus \, \, INV}$ & 2.89 & 71.3  &  57.7 & 81.5& 96.4 & 2.98 & 83.8 & 74.8  & 91.9 & 96.5 \\
%   \cline{1-11}
   $\fol{AUG\, \,  minus \, \, FIL}$ & 2.84  & 72.5 & 59.5  & 82.3  & 96.4 & 3.01 & 83.9  & 75.1 & 91.5  & 96.5  \\
%   \cline{1-11}
   $\fol{AUG\, \,  minus \, \, RW}$ &  2.99 & 70.7  & 57.1  & 80.8 & 
   95.6 & 3.05 & 82.8 & 73.2 & 91.1 & 96.5\\
%\Xhline{3\arrayrulewidth}
\Xhline{3\arrayrulewidth}
\end{tabular}
\label{tab:AppendixAblationStudyKinshipAndUMLS}
\vspace{0.12in}

% FOURTH TABLE ENDS

%FIFTH TABLE STARTS
\centering
\small
\caption{Ablation study performed on WN18RR for abduction ($\fol{ABD}$), inversion ($\fol{INV}$), filtering ($\fol{FIL}$) and PCA-based random walk augmentation ($\fol{RW}$). $\fol{AUG}$ represents represents all the approaches proposed in our work.}
\vspace{0.03in}
\begin{tabular}{|c|ccccc|} 
\Xhline{3\arrayrulewidth}
%\Xhline{3\arrayrulewidth} 
\multirow{2}{4em}{\textbf{Algorithm}} & 
\multicolumn{5}{c|}{\textbf{WN18RR}}  \\
\cline{2-6}
& MR & MRR & H@1 & H@3 & H@10  \\
%\Xhline{3\arrayrulewidth}
\Xhline{3\arrayrulewidth}
   $\fol{AUG }$ & 4231.77  & \textbf{55.0} & \textbf{51.0}  & \textbf{57.2} & \textbf{63.5}  \\
 %  \cline{1-6}
   $\fol{ AUG \, \,  minus \, \,  ABD }$ & 4406.95  & 52.2 & 47.8 & 54.1 & 61.0 \\
 %  \cline{1-6}
   $\fol{AUG \, \,  minus \, \, INV}$ & 4302.04  & 54.4  & 50.0 & 56.8 & 62.7 \\
 %  \cline{1-6}
 $\fol{AUG\, \,  minus \, \, FIL}$ &  \textbf{4224.20} & \textbf{55.0}  & 50.6  & 57.1 & 63.3 \\
 %  \cline{1-6}
 $\fol{AUG\, \,  minus \, \, RW}$ & 4263.43  & 54.6  & 50.1  & 57.0 & 63.2\\
%\Xhline{3\arrayrulewidth}
\Xhline{3\arrayrulewidth}
\end{tabular}
\label{tab:appendixAblationStudyWN18RR}
\vspace{0.1in}
% FIFTH TABLE ENDS
\end{table*} % MAIN TABLE ENDS

\section{Data Statistics and Evaluation Metrics}
\label{appendix: datastatistics}
Table \ref{tab:KG statistics} summarizes the statistics of the data used in the experiments of our work. We utilize the standard train, validation and test splits for WN18RR and FB15k-237 datasets. Since there are no standard splits for UMLS and Kinship datasets, for consistency, we employ the splits used by RNNLogic~(\citeyear{RNNLogic2021}) for evaluation (created by randomly sampling 30\% triplets for training, 20\% for validation and the rest 50\% for testing). 

\paragraph{Metrics:} For each triplet $\fol{(h, r, t)}$ in the test set, traditionally queries of the form $\fol{(h, r, ?)}$ and $\fol{(?, r, t)}$ are created for evaluation, with answers $\fol{t}$ and $\fol{h}$ respectively. We model the $\fol{(?, r, t)}$ query as $\fol{(t, r^{-1}, ?)}$ with the same answer $\fol{h}$, where $\fol{r^{-1}}$ is the inverse relation for $\fol{r}$. In order to train the model over the inverse relations, we augment the training data with an additional $\fol{(t, r^{-1}, h)}$ triple for every triple $\fol{(h, r, t)}$ present in KG.

Given ranks for all queries, we report the Mean Reciprocal Rank (MRR) and Hit@k (H@k, k = 1, 10) under the filtered setting in the main paper and two additional metrics: Mean Rank (MR) and Hits@3 in the appendices. MRR and Hits@k metrics are reported after multiplying with 100. To maintain consistency with RNNLogic, in cases where the model assigns same probability to other entities along with the answer, we compute the rank as $(m + \frac{(n+1)}{2})$ where $m$ is the number of entities with higher probabilities than the correct answer and $n$ is the number of entities with same probability as the answer.

\section{RotatE}
\label{appendix: rotate}
$\fol{RotatE}$ is a knowledge graph embedding model that embeds entities and relations in complex space. Relation embeddings are modeled as rotations in complex vector space. Formally, $\fol{RotatE} (\fol{h}, \fol{r}, \fol{t})$ is calculated using the following equation:
\begin{equation}
\label{eq: RotatE Score}
\fol{RotatE}\,(\fol{h}, \fol{r}, \fol{t}) = -\fol{d} (\fol{\textbf{x}}_{\fol{h}} \circ \fol{\textbf{x}}_{\fol{r}}, \fol{\textbf{x}}_{\fol{t}})
\end{equation}
where $\fol{d}$ is the cosine distance in complex vector space, RotatE embedding of $\fol{r}$ is $\fol{\textbf{x}_{r}}$, and $\circ$ is the Hadamard product. Intuitively, we rotate $\fol{\textbf{x}}_h$ by the rotation defined by $\fol{\textbf{x}}_r$ and consider the distance between the result and $\fol{\textbf{x}}_t$. For our experiments, $\fol{RotatE}$ is trained separately and the trained embeddings are used to calculate scores for the [$\fol{RNN + RotE}$] baseline.

 \section{Experimental Setup for RNNLogic}
 \label{sec:rulestatistics}
In order to obtain main results in Table \ref{tab:tablemainMRRAndHitsAt10}, we train the rule generator in RNNLogic with optimal hyperparameters obtained after communication with the original authors and generate a set of high-quality Horn rules to use for training RNNLogic+. 
For our best results, we utilize optimal rules provided by the authors of 
RNNLogic\footnote{\url{https://github.com/DeepGraphLearning/RNNLogic}}. % filter ($\fol{FIL}$) these rules by PCA score and further
We augment these rules by abduction ($\fol{ABD}$), and then rule inversion ($\fol{INV}$) on both the original rules and the rules formed after abduction. We further augment the rulebase with the rules discovered by random walks ($\fol{RW}$). Finally, we filter ($\fol{FIL}$) superior rules from these rules by PCA score. We present statistics detailing the number of rules used per dataset after each augmentation step in Table \ref{tab:rule size}. These rules are utilized in RNNogic+ ($\fol{[RNN]}$-$\fol{(Orig)}$) and RNNLogic+ with RotatE ($\fol{[RNN}$+$\fol{RotE]}$-$\fol{(Orig)}$) baselines. For the other results: $\fol{[RNN]}$-$\fol{(RW)}$ and $\fol{[RNN}$+$\fol{RotE]}$-$\fol{(RW)}$, we employ only the rules obtained by $\fol{RW}$ augmentation and train RNNLogic+ model with them (Appendix \ref{appendix:rulegenerationviarandomwalks}). The goal of these set of results is to test the utility of abduction and rule inversion with a different set of rules. 
The details of training RNNLogic+ model is provided in Appendix \ref{appendix:RNNLogic+training}.

\section{RNNLogic Results Reproduction}
\label{appendix: rnnlogicresultsreproduction}
We have reproduced the results of RNNLogic+ with and without RotatE and obtained similar results to  the original RNNLogic paper~\cite{RNNLogic2021}, however the numbers reported in this paper for $\fol{[RNN]}$ and $\fol{[RNN + RotE]}$ are our own reproductions. In this section, we report a comparison between the original results and our reproduced results for $\fol{RNNLogic+}$ model ($\fol{[RNN]}$) on the WN18RR and FB15K-237 datasets. As can be observed in the Table \ref{tab:originalvsreproduced}, our reproduced results are better than the published results of RNNLogic model for both the datasets, using hyperparameters obtained after communication with the authors of the RNNLogic paper.

\begin{table}[H]
\centering
\small
\caption{Comparison of the results reported in original RNNLogic paper with the results reproduced by the authors of this paper.}
\label{tab:originalvsreproduced}
\begin{center}
\resizebox{\columnwidth}{!}{
\begin{tabular}{|p{1.2cm}p{1.3cm} p{0.5cm}p{0.6cm}p{0.6cm}p{0.6cm}p{0.75cm}|} 
 \cline{1-7}
\textbf{Dataset} &	\textbf{Numbers} &	\textbf{MR} &	\textbf{MRR} &	\textbf{H@1} & \textbf{H@3} &	\textbf{H@10} \\
 \cline{1-7}
\multirow{2}{*}{WN18RR} & Reported &	7204 & 48.9 & 45.3 & 50.6 &	56.3 \\
& Reproduced & \textbf{5858} & \textbf{49.6} & \textbf{45.5} & \textbf{51.4} & \textbf{57.4} \\
 \cline{1-7}
 \multirow{2}{*}{FB15K-237} & Reported & 480 &	29.9 & 21.5 & 32.8 & 46.4\\
 & Reproduced & \textbf{256} & \textbf{32.9} &	\textbf{24.0} &	\textbf{36.1} &	\textbf{50.6} \\
  \cline{1-7}
\end{tabular}}
\end{center}
\vspace{-0.20in}
\end{table}

\section{ExpressGNN Training and Hyperparameter Setting}
\label{appendix: expressGNNbaseline}
As already discussed in Section \ref{sec:Experiments}, in order to prove the broad applicability of proposed augmentations ($\fol{AUG}$) in our work, we perform experiments with ExpressGNN model as another baseline in Table \ref{tab:expressGNNresults}. In this section we provide the details of this experiment. The current implementation of $\fol{ExpressGNN}$ model scales poorly with the number of rules, necessitating the use of a much smaller ruleset size. We generate ruleset for each dataset by selecting the top $5-10$ rules per relation (in the rule head) from RNNLogic rules for that dataset ($\fol{ORIG}$) based on the PCA score. This results in 417 rules for WN18RR, 500 rules for Kinship and 460 rules for UMLS. We perform augmentations on these rules and further maintain a threshold of the PCA score to be 0.95 while filtering $\fol{RW}$ rules. After augmentation, we obtain 1734 rules for Kinship, 2058 rules for UMLS and 828 rules for WN18RR. We also augment the training and the test set of ExpressGNN datasets with the inverse triples ($\fol{t}, \fol{r}^{-1}, \fol{h}$) for each original ($\fol{h}, \fol{r}, \fol{t}$) triple. Hyperparameters used for training are the optimal ones from the original paper. Results for FB15k-237 are omitted since ExpressGNN does not scale up to the augmented ruleset.

ExpressGNN assumes the knowledge of test queries at training time to construct its Markov Logic Network. For the test triple ($\fol{h}, \fol{r}, \fol{t}$), this informs the model that $\fol{h}$ is a potential head and $\fol{t}$ is a potential tail entity for given relation $\fol{r}$, even though this information might not be present in the training data. Hence, results presented in Table \ref{tab:expressGNNresults} are not directly comparable to results in Table \ref{tab:tablemainMRRAndHitsAt10}.

\section{Rule Generation via Random Walks}
\label{appendix:rulegenerationviarandomwalks}
Because rules generated by employing random walks form a distinct ruleset in the main paper ($\fol{[RNN]}$-$\fol{(RW)}$), we explain the statistics of these rules in detail in a dedicated section here. In order to determine the number of rules generated from the random walks, we calculate the difference of the column `\textbf{\#Rules} + $\fol{INV}$ + $\fol{ABD}$' and `\textbf{\#Rules} + $\fol{INV}$ + $\fol{ABD}$ + $\fol{RW}$' in the Table \ref{tab:rule size} and summarize the resulting statistics of the number of $\fol{RW}$ rules created for each dataset in the Table  \ref{tab:randomwalkrulesperdataset}. When compared to Table \ref{tab:rule size}, we note that although random walk rules ($\fol{RW}$) comprise less than 8\% of the augmented ruleset for all the datasets, these rules are still pivotal. This is because we notice a considerable decrease in performance after removing these rules as observed in Table \ref{tab:tableAblationStudy}, Table \ref{tab:AppendixAblationStudyKinshipAndUMLS} and Table \ref{tab:appendixAblationStudyWN18RR}.

\begin{table}[h]
\caption{Number of random walk rules ($\fol{RW}$) generated per dataset in the experiments}
\label{tab:randomwalkrulesperdataset}
\centering
\small
\begin{center}
\resizebox{\columnwidth}{!}{
%\begin{tabular}{|p{1.7cm} p{1.5cm}|} 
\begin{tabular}{|p{1.5cm}|p{1.52cm}|p{1.1cm}| p{0.9cm}|p{0.9cm}|} 
%\cline{1-2}
%\textbf{Dataset} &	\textbf{\#RW Rules} \\
%\cline{1-2}
%FB15K-237 &	2687\\
%WN18RR & 2425 \\
%Kinship	& 14219 \\
%UMLS &	10313\\
\cline{1-5}
\textbf{Dataset} & FB15K-237 & WN18RR & Kinship & UMLS \\
\cline{1-5}
\textbf{\#RW Rules} & 2687 & 2425 & 14219 & 10313 \\
\cline{1-5}
\end{tabular}
}
\end{center}
\vspace{-0.15in}
\end{table}

\section{RNNLogic+ Training and Hyperparameter Setting}
\label{appendix:RNNLogic+training}
Here we describe the training of RNNLogic+ model that is utilized in Table \ref{tab:tablemainMRRAndHitsAt10} and complementary Table \ref{tab:appendixMainDetailedresults}. We use the same methodology for training RNNLogic+ model as in the original work~\cite{RNNLogic2021}. New rule embeddings are created for all the rules that are added to the rule set after rule augmentation. Rule embedding dimension is set to $16$ (compared to $32$ in original RNNLogic+) across datasets to mitigate the effect of the increased number of parameters in the model due to new rule embeddings. Results reported are for a single run with fixed seed over $5$ epochs of training.

The hyperparameter $\eta$ in Equation (\ref{eq: RNNlogicplus and KGE score}) representing the relative weight is set to 0.01, 0.05, 0.1 and 0.5 for WN18RR, FB15k-237, UMLS and Kinship respectively. The RotatE embedding dimension is set to 200, 500, 1000 and 2000 for WN18RR, FB15k-237, UMLS and Kinship respectively. We keep a consistent batch size of 8, 4, 32 and 16 for WN18RR, FB15k-237, UMLS and Kinship respectively. The number of parameters for RNNLogic+ scales with the rule embedding size and the number of rules, reaching a maximum of 16*298446 = 4775136 for FB15k-237 after augmentations and filtering (leading to a training time of around 23 hours). As we can see, augmentation adds new rules leading to increase in the parameters of the model. All training was carried out on a single Tesla V100 GPU.    %\nk{Ananjan, explain the hperparameters of Kinship here} 
The optimal values of all the hyper-parameters was found by tuning on validation set on each dataset.
\section{Detailed Results on Proposed Augmentations}
\label{appendix:RNNLogic+results}
Results in Table \ref{tab:appendixMainDetailedresults} are supplementary to results already presented in Table \ref{tab:tablemainMRRAndHitsAt10}. 
%In Table $\fol{RWRL}$ constitute a model whose rules are generated by performing random walks on a KG followed by filtering high-quality rules utilizing PCA-score. These rules are further input to RNNLogic+ model for training. $\fol{RWRL}$-$\fol{RotE}$ is the $\fol{RWRL}$ model that is ensembled with RotatE model, analogous to equation \ref{eq: RNNlogicplus and KGE score}.
In addition to MRR, Hits@1 and Hits@10 presented in the Table \ref{tab:tablemainMRRAndHitsAt10} in the Experiment section, we also present Mean Rank (MR) and Hits@3 here. As discussed already in Section \ref{sec:Experiments}, $\fol{AUG}$ includes abduction ($\fol{ABD}$), inversion ($\fol{INV}$), rule filtering ($\fol{FIL}$) and random walk augmentation ($\fol{RW}$).

In Table \ref{tab:appendixMainDetailedresults}, we observe that there is a consistent improvement in the performance of the model for all the metrics after rule augmentation and filtering ($\fol{AUG}$). Notably, for the two new metrics introduced in Table \ref{tab:appendixMainDetailedresults}, we obtain a performance gain of 3.7 point on Hits@3 and 40.4\% on MR for FB15K-237 dataset and
$\fol{[RNN]}$-$\fol{(RW)}$  baseline. Since the original rules for the random walk baseline are lesser in number, $\fol{[RNN]}$-$\fol{(RW)}$ and $\fol{[RNN}$ + $\fol{RotE]}$ - $\fol{(RW)}$ benefit more from augmentation. We also observe that for Kinship and UMLS, $\fol{[RNN}$ + $\fol{RotE]}$ - $\fol{(RW)}$ gives better performance than $\fol{[RNN}$ + $\fol{RotE]}$ - $\fol{(Orig)}$, highlighting the quality of the rules discovered by local random walks followed by PCA filtering. 
%We claim that the reason for significant improvement is that our rule augmentation techniques result in better utilization of the learned high-quality rules. The groundings from the new rules can be used to create new paths between entities in the Knowledge Graph, resulting in more informed scores as computed by RNNLogic+ model.
\section{Detailed Results of Ablation Study}
\label{appendix:RNNLogic+ablation}
Results in Table \ref{tab:AppendixAblationStudyKinshipAndUMLS} are supplementary to results already presented in Table \ref{tab:tableAblationStudy}. Besides the three metrics presented in Table \ref{tab:tableAblationStudy}, we present Hits@3 and MR in this table. Additionally, we also demonstrate results of ablation on WN18RR dataset in Table \ref{tab:appendixAblationStudyWN18RR}. Ablation is not performed on FB15K-237 due to computational constraints. As with the other metrics, Hits@3 and MR is the most affected  by abductive rules in UMLS and WN18RR because abduction results in augmenting the ruleset with a large number of high-quality rules (see Table \ref{tab:tableFOILandPCA}). Furthermore, Hits@3 and MR gets most affected by PCA-based random walk augmentation in Kinship dataset. This is because Kinship is a dense dataset, and a large number of high-quality rules are quickly discovered by the random walks.

\section{Detailed Results of Rule Generation vs Rule Augmentation}
\label{appendix:RuleAugmentationvsRuleGeneration}
\begin{table*}[t]
 \caption{Comparison of performance by rule augmentation with performance on the original rules on WN18RR and Kinship. R/R and TR is number of rules per relation and total rules generated from RNNLogic respectively. ABD represents abduction performed on original rules.}
  \vspace{-0.03in}
 \label{tab:AppendixAugumentationvsgeneration}
 \centering
\small
\begin{center}
\begin{tabular}{|p{1.20cm}|cccccccc|} 
\Xhline{3\arrayrulewidth}
%\Xhline{3\arrayrulewidth} 
Dataset & R/R& TR & ABD & MR & MRR & Hits@1 & Hits@3& Hits@10 \\
\Xhline{3\arrayrulewidth}
%\Xhline{3\arrayrulewidth}
\multirow{2}{4em}{WN18RR} & 80 & 9867 & Yes & \textbf{4701.61}& \textbf{49.0}  & \textbf{44.9}  & \textbf{50.5} & \textbf{56.7}  \\
%\cline{2-9}
& 500 & 11000 & No & 4848.39 & 47.7 & 43.7 & 49.8 & 55.2  \\
\Xhline{3\arrayrulewidth} 
\multirow{2}{4em}{Kinship} &80 & 18432 & Yes & \textbf{3.21} & \textbf{69.5} & \textbf{56.1} & \textbf{79.4} & \textbf{94.6}  \\
%\cline{2-9}
 & 500 & 25000 & No & 3.62 & 66.1 &52.1 & 75.3 & 93.1 \\
%\Xhline{3\arrayrulewidth}
\Xhline{3\arrayrulewidth}
\end{tabular}
\end{center}
\vspace{-0.25in}
\end{table*}

Results in Table \ref{tab:AppendixAugumentationvsgeneration} are supplementary to the results already presented in Table \ref{tab:tableAugumentationvsgeneration}. Here we present Hits@3 and MR as two additional metrics for analyzing the need for rule augmentation. 

We generate rules by training RNNLogic model. We consider 80 rules per relation for each dataset from these rules and expand them by performing three augmentations and filtering. This results in total of $9867$ rules for WN18RR and $18432$ rules for Kinship data. Then, we train RNNLogic+ with RotatE ($\fol{[RNN}$+$\fol{RotE]}$) on these rules and compare the results with RNNLogic+ with RotatE model trained on 500 rules per relation without augmentations. We observe that model trained with augmented rules consistently performs better than model trained by merely increasing the number of rules generated, even for a comparable number of rules. Specifically, we observe that model trained with augmented rules shows 4 point Hit@1 gain in Kinship dataset over the model trained with merely increased rules.  These results strengthens the hypothesis that it is more helpful to leverage a few high-quality augmented rules rather than exploiting a plethora of lower-quality rules for Neuro-Symbolic KG Completion.

\section{Qualitative Analysis of the\\ Augmented Rules}
\label{appendix:qualitativeanalysisofrules}

In this section, we present one logical rule generated after each augmentation step as examples. The rules are taken from the FB15K-237 dataset.
\begin{enumerate}
\item \textbf{ABD}: $\fol{LivesIn(PersonA, LocationB)}:-$  $\fol{PlayFor(PersonA, TeamC)}$, $\fol{Inverse\_Team}$ $\fol{\_Location(TeamsC, LocationB)}$

\item \textbf{INV}: $\fol{Inverse\_Person\_Language(Langu}$ $\fol{ageA, PersonB) :-\, Inverse\_Film's\_Lang}$ $\fol{uage(LanguageA, FilmC),\, StoryWritten}$ $\fol{By(FilmC, PersonB)}$

\item \textbf{RW}: $\fol{Friends(PersonA, \, PersonB)}$ $:-$  $\fol{Friends(PersonA, \, PersonC), Inverse\_}$ $\fol{Producer(PersonC, \, FilmD), Writer(FilmD}$ $\fol{,\, PersonB)}$
\end{enumerate}

\noindent For example, the rule in \textbf{ABD} category states that a person will live in the same city as the team he plays for is located. Therefore, we conclude that the rules captured through augmentations can be human interpretable.

\section{An Alternative Augmentation Strategy}
\label{appendix: augmentationsonOrigandRw}
Recall that in our proposed methodology ($\fol{Orig}$+$\fol{AUG}$) in Section \ref{sec:main}, we consider original rules ($\fol{ORIG}$) and perform abduction ($\fol{ABD}$) on the original rules. This is followed by rule inversion ($\fol{INV}$) over the original rules and abductive rules. Then, we introduce the random walk rules ($\fol{RW}$) as the final augmentation step in the proposed augmentations ($\fol{AUG}$) for the original ($\fol{Orig}$) ruleset. In this section, we consider an alternative sequence of augmenting the ruleset where we consider both the original ($\fol{Orig}$) and the random walk rules ($\fol{RW}$) and apply abduction and rule inversion on both of them. We denote this setting as $\fol{(Orig + AUG2)}$.  We report a comparison of $\fol{(Orig + AUG2)}$ with $\fol{(Orig + AUG)}$ (Table \ref{tab:tablemainMRRAndHitsAt10}) with $\fol{[RNN + RotE]}$ as the baseline model in Table \ref{tab:origandaug2setting}. From the results in the table, we conclude that $\fol{Orig}$ + $\fol{AUG2}$ does not result in improvement over our original methodology of $\fol{Orig}$ + $\fol{AUG}$. %\nk{Ananjan, do we know the reason, why?} [Not Really: Ananjan] 
It also creates a larger ruleset, further slowing down the training of the model.

\begin{table}[h]
\caption{Comparison of performance by exploring two methodologies of augmentations: $\fol{(Orig + AUG)}$ and $\fol{(Orig + AUG2)}$.}
\label{tab:origandaug2setting}
\centering
\small
\begin{center}
\begin{tabular}{|p{1.4cm}p{1.7cm} p{0.8cm}p{0.8cm}p{0.8cm}|} 
 \cline{1-5}
\textbf{Dataset} &\textbf{Augmentation} & \textbf{MRR} & \textbf{H@1} & \textbf{H@10} \\
 \cline{1-5}
\multirow{2}{4em}{\textbf{WN18RR}} & $\fol{Orig}$ + $\fol{AUG}$	& \textbf{55.0} &	\textbf{51.0} &	\textbf{63.5}	\\
& $\fol{Orig}$ + $\fol{AUG2}$ &	54.4 &	50.2 &	62.9 \\
 \cline{1-5}
\multirow{2}{4em}{\textbf{Kinship}} & $\fol{Orig}$ + $\fol{AUG}$ &	\textbf{72.9} &	\textbf{59.9} & \textbf{96.4} \\
& $\fol{Orig}$ + $\fol{AUG2}$ &	71.1 &	58 & 95.8 \\
\cline{1-5}
\end{tabular}
\end{center}
\vspace{-0.20in}
\end{table}

\section{PCA-Confidence Metric}
\label{appendix:PCAScoringFunction}

In this section, we explain in detail, the PCA-confidence metric that has been employed to score the rules discovered through random walk in our third augmentation approach. This metric has also been used to score the augmented rules in Table \ref{tab:tableFOILandPCA}. 

\vspace{0.5ex}
\noindent \textbf{PCA}: The calculation of the metric utilizes a Partial Closed World assumption~\cite{AMIE2013} and assumes that if we know one $\fol{t}$ for a given $\fol{r}$ and $\fol{h}$ in $\fol{r(h, t)}$, then we know all $\fol{t'}$ for that $\fol{h}$ and $\fol{r}$. Let the rules under consideration be of the form $\fol{B} \Rightarrow \fol{r}(\fol{h}, \fol{t})$. Then the PCA-score $\fol{PCAConf(B \Rightarrow r)}$ is:
\scalebox{0.9}{\parbox{.5\linewidth}{%   
\begin{align*}
    \label{eq: PCA Score}  
     \fol{\dfrac{\#(h,t) :  |Path(h, B, t)| > 0 \wedge r(h,t) \in P}{\#(h,t) : |Path(h, B, t)| > 0 \wedge \exists t': r(h,t') \in P}}
\end{align*}
}}

Essentially, it is the number of positive examples, $\fol{P}$, satisfied by the rule divided by the total number of $\fol{(h,t)}$ satisfied by the rule such that $\fol{r(h,t')}$ is a positive example for some $\fol{t'}$. 

\section{FOIL-Score Metric}
\label{appendix:FOILScoringFunction}
 We employ a modification of FOIL as one of the evaluation metrics to assess the quality of rules produced by augmentation techniques (\textbf{Q1}) in Table \ref{tab:tableFOILandPCA}. FOIL-scoring metric is discussed in detail below. 
\noindent \textbf{FOIL:} Let the rules be of the form $\fol{B} \Rightarrow \fol{r}(\fol{h}, \fol{t})$. Let $\fol{Path(h, B, t)}$ be the set of paths from $\fol{h}$ to $\fol{t}$ that act as groundings for the rule body $\fol{B}$. Under the Closed World assumption, we assume that all triples not in the training and test set are false. Inspired by the First-Order Inductive Learner algorithm~\cite{FOIL1990}, we define FOIL score to assess the quality of a rule as follows:
    \begin{equation*}
    \label{eq: FOIL Score}
    \fol{FOIL(B \Rightarrow r) = \frac{\sum_{r(h, t) \in P}|Path(h, B, t)|}{\sum_{(h,t)}|Path(h, B, t)|}}
    \end{equation*}

In the above equation, $\fol{P}$ represents the set of positive examples in the given KG. The key difference between the FOIL score proposed originally~\cite{FOIL1990} and ours is that instead of considering the number of examples satisfied by the rule, we calculate the number of groundings of the rule. This is more in line with the score calculated by RNNLogic+, which considers the number of groundings as well. Ideally the rules should have larger number of groundings for positive triples in comparison to the other triples. 

Typically, negative sampling is used to calculate these metrics (PCA in Appendix \ref{appendix:PCAScoringFunction} and FOIL here) as it is computationally expensive to compute exhaustive negative examples. However, we calculate these metrics by considering the entire knowledge graph, which is enabled by utilizing batching and sparse matrix operations on the adjacency graph. 

We highlight that we are the first to show the utility of PCA Confidence and FOIL in the context of neuro-symbolic models. This makes our specific approach distinct from AMIE~\cite{AMIE2013} and FOIL~\cite{FOIL1990}, and more targeted to our setting due to the changes in the method of computation.

%\begin{table}[H]
% \caption{Comparison of performance after augmentation with performance on rules from generator on WN18RR and Kinship. R/R and TR is number of rules per relation and total rules generated from RNNLogic respectively.}
%  \vspace{-0.07in}
% \label{tab:AppendixAugumentationvsgeneration}
% \centering
%\small
%\begin{center}
%\begin{tabular}{|c?c|c?c|c?} 
%\Xhline{3\arrayrulewidth}
%\textbf{Metric} & \multicolumn{2}{c?}{\textbf{WN18RR}} & \multicolumn{2}{c?}{\textbf{Kinship}} \\
%\Xhline{3\arrayrulewidth}
%R/R & 100 & 500 & 100 & 500 \\
%\hline
%TR & 11583 & 11000 & 26837 & 25000 \\
%\hline
%AUG & Yes & No & Yes & No \\
%\hline
%MR & 4634.80 & 4848.39 & 3.08 & 3.62 \\
%\hline
%MRR & 49.5 & 47.7 & 70.4 & 66.1 \\
%\hline
%Hits@1 & 45.5 & 43.7 & 57.1 & 52.1 \\
%\hline
%Hits@3 & 51.0 & 49.8 & 80.1 & 75.3 \\
%\hline
%Hits@10 & 57.4 & 55.2 & 95.4 & 93.1 \\
%\Xhline{3\arrayrulewidth}
%\Xhline{3\arrayrulewidth}
%\end{tabular}
%\end{center}
%\vspace{-0.25in}
%\end{table}

\end{document}